\documentclass[manuscript]{acmart} 

\renewcommand\footnotetextcopyrightpermission[1]{}
\usepackage{microtype}
\usepackage{xurl}
\usepackage{graphicx}
\usepackage{subcaption}
\usepackage{overpic}
\usepackage{adjustbox}
\usepackage{float}
\usepackage{booktabs}
\usepackage{array}
\usepackage{multirow}
\usepackage{longtable}
\usepackage{amsmath} 
\usepackage{enumitem}
\setlist{nosep}
\usepackage{xcolor}
\usepackage{verbatim}
\usepackage{textcomp}
\usepackage{algorithm}
\usepackage{algpseudocode}
\usepackage{balance}

\AtBeginDocument{
  \setlength{\abovedisplayskip}{6pt plus 2pt minus 2pt}
  \setlength{\belowdisplayskip}{6pt plus 2pt minus 2pt}
  \setlength{\abovedisplayshortskip}{4pt plus 2pt minus 2pt}
  \setlength{\belowdisplayshortskip}{4pt plus 2pt minus 2pt}
}
\begin{document}

\title{Detection and Recovery of Adversarial Slow-Pose Drift in Offloaded Visual-Inertial Odometry}

\author{Sourya Saha}
\authornote{Both authors contributed equally to this work.}
\affiliation{%
  \institution{The Graduate Center, City University of New York}
  \city{New York}
  \state{NY}
  \country{USA}
}
\email{ssaha2@gradcenter.cuny.edu}

\author{Md Nurul Absur}
\authornotemark[1] 
\affiliation{%
  \institution{The Graduate Center, City University of New York}
  \city{New York}
  \state{NY}
  \country{USA}
}
\email{mabsur@gradcenter.cuny.edu}

\author{Saptarshi Debroy}
\affiliation{%
  \institution{Hunter College and The Graduate Center, City University of New York}
  \city{New York}
  \state{NY}
  \country{USA}
}
\email{saptarshi.debroy@hunter.cuny.edu}

\begin{abstract}

Visual–Inertial Odometry (VIO) supports immersive Virtual Reality (VR) by fusing camera and Inertial Measurement Unit (IMU) data for real-time pose. However, current trend of offloading VIO to edge servers can lead server-side threat surface where subtle pose spoofing can accumulate into substantial drift, while evading heuristic checks. 
In this paper, we study this threat and present an unsupervised, label-free detection and recovery mechanism. The proposed model is trained on attack-free sessions to learn temporal regularities of motion to detect runtime deviations and initiate recovery to restore pose consistency. We evaluate the approach in a realistic offloaded-VIO environment using ILLIXR testbed across multiple spoofing intensities. Experimental results in terms of well-known performance metrics
show substantial reductions in trajectory and pose error compared to a no-defense baseline. 
\end{abstract}



\acmYear{2025}\copyrightyear{2025}
\setcopyright{cc}
\setcctype[4.0]{by}
\acmConference[MobiHoc '25]{The Twenty-sixth International Symposium on Theory, Algorithmic Foundations, and Protocol Design for Mobile Networks and Mobile Computing}{October 27--30, 2025}{Houston, TX, USA}
\acmBooktitle{The Twenty-sixth International Symposium on Theory, Algorithmic Foundations, and Protocol Design for Mobile Networks and Mobile Computing (MobiHoc '25), October 27--30, 2025, Houston, TX, USA}
\acmDOI{10.1145/3704413.3765307}
\acmISBN{979-8-4007-1353-8/25/10}

\maketitle

\section{Introduction}
\label{sec:intro}

Virtual Reality (VR) systems are expanding across entertainment, education, and industry, enabled by lightweight headsets and high-fidelity rendering. At the core of these experiences lies Visual-Inertial Odometry (VIO) \cite{8593941}, which fuses camera and Inertial Measurement Unit (IMU) data to estimate six-Degrees of Freedom (DoF) head pose. To mitigate the power and thermal limits of mobile headsets, recent designs offload such VIO computation to nearby edge servers to preserve resource-constrained devices and extend battery life \cite{10.1145/3712676.3714444}. Although such offloading improves end-to-end performance, it opens a new threat surface where server-returned poses may get delayed, degraded, or adversarially manipulated before reaching the headset.

\begin{figure}[t]
    \centering
    \begin{subfigure}[t]{0.47\linewidth}
        \includegraphics[width=\linewidth]{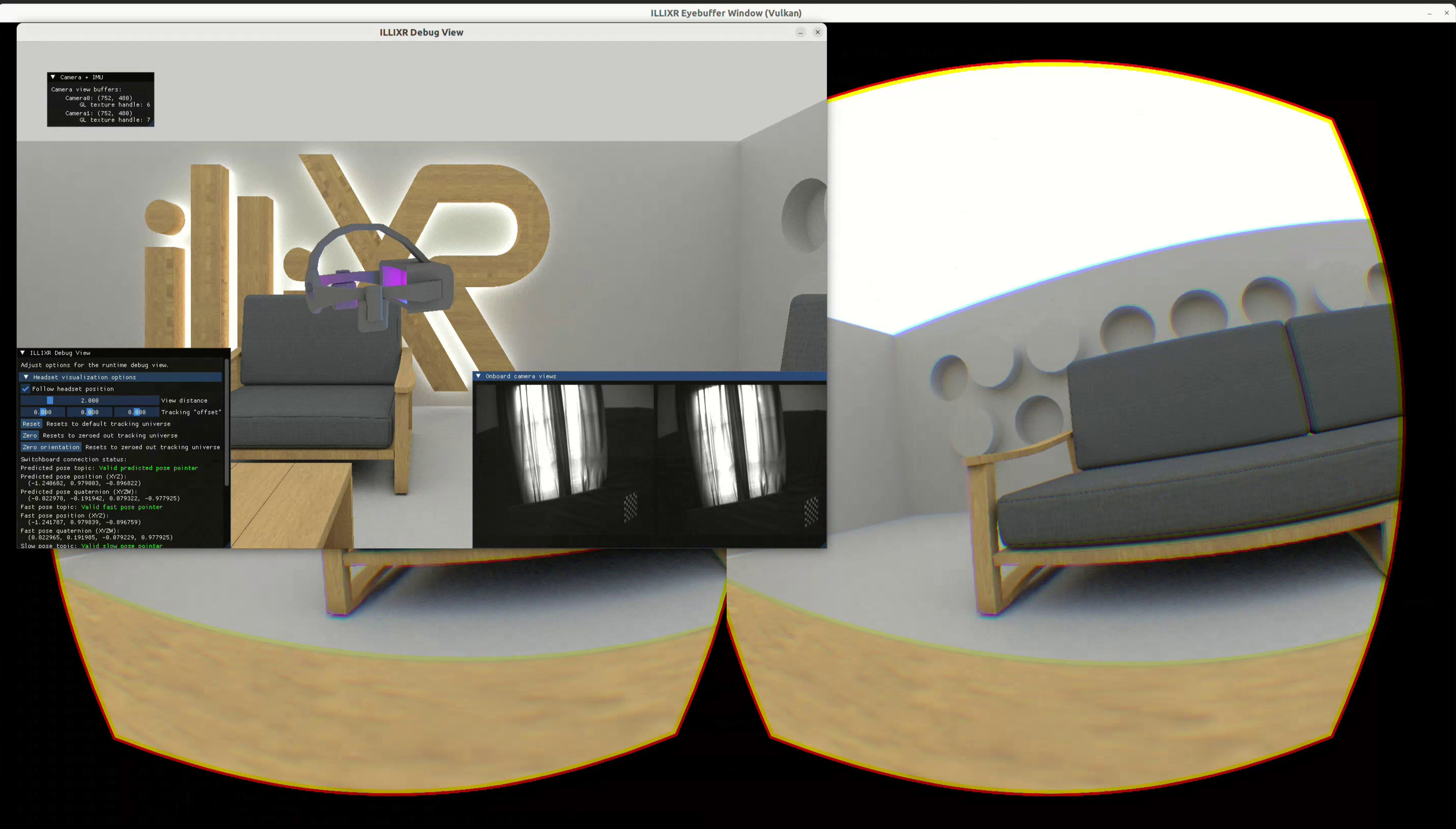}
        \caption{No pose spoofing}
    \end{subfigure}
    \hspace{1em} 
    \begin{subfigure}[t]{0.47\linewidth}
        \includegraphics[width=\linewidth]{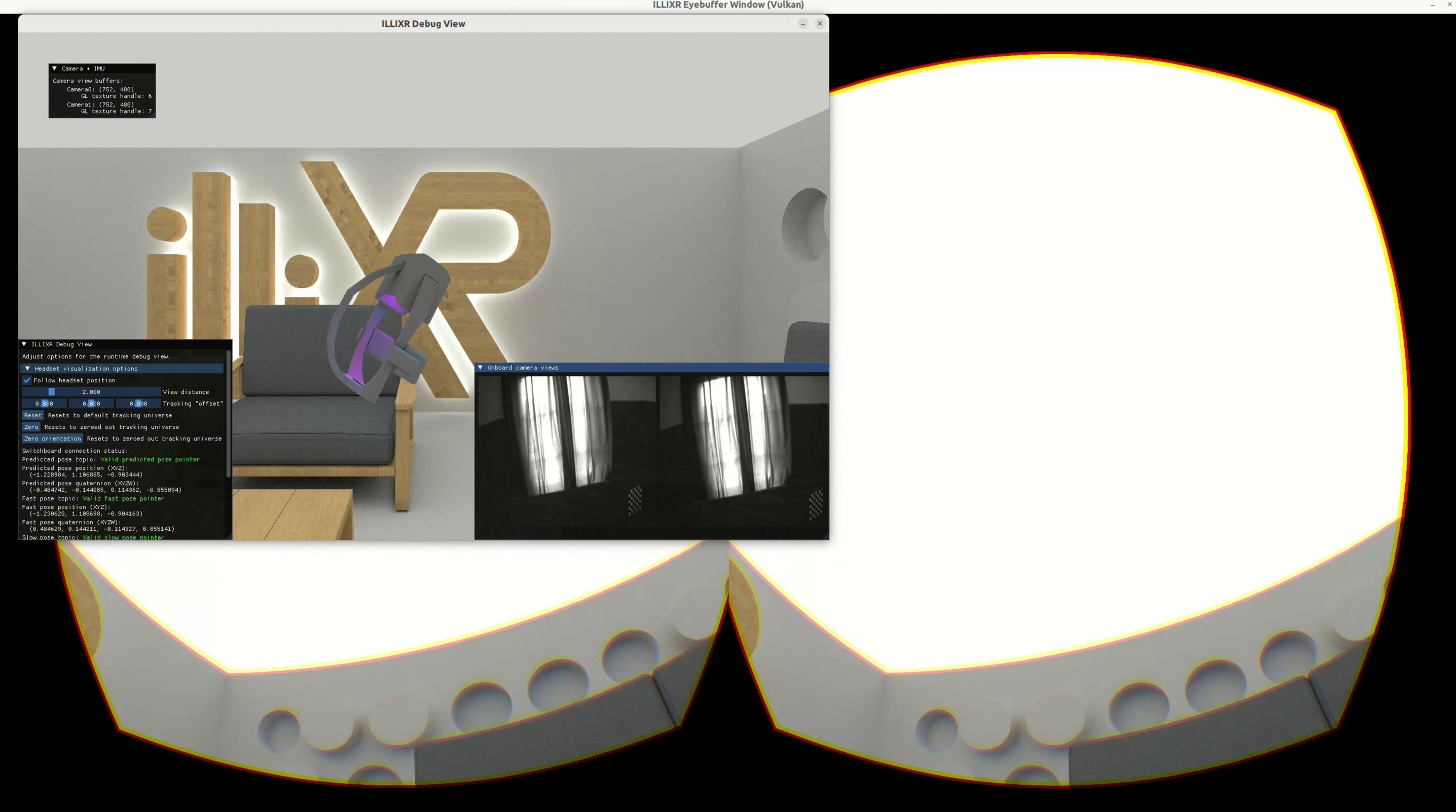}
        \caption{Significant spoofing}
    \end{subfigure}
    \caption{QoE impact of subtle VIO perturbations. Same head pose: normal (left) vs. anomalous (right).}
    \label{fig:pose-manipulation}
    \vspace{-10pt} 
\end{figure}

One of such subtle yet damaging vector to manipulate such server-returned poses can be \emph{adversarial slow-pose drift} which are low-magnitude, temporally consistent perturbations injected into the offloaded head-pose stream. Such drifts can evade residual filters and conventional anomaly detectors, but compound over time into spatial misalignment, visual jitter, and degraded immersion (Figure~\ref{fig:pose-manipulation}). Existing edge-offloaded VIO pipelines typically assume a trustworthy server and lack explicit verification, including commercial and research frameworks such as ARCore \cite{10348201}, HoloLens \cite{s22134915}, and ILLIXR \cite{illixr}, leaving them exposed to continuous, low-rate manipulation \cite{xu2023sok, 
wetterstrom2023virtual}.
Prior efforts to mitigate pose drift in VIO/SLAM include loop closures \cite{9780121}, pose-graph optimization \cite{9425438}, Kalman-filtering variants \cite{10411891}, residual-based thresholding \cite{10495730, PEREIRA2023118884}, predictive delay compensation, and adversarial filtering for SLAM \cite{9543603,yoshida2022adversarial, LI2023105344}. {\em These methods can be effective in high-motion scenes or when deviations are large and easily flagged. However, they are less suited to slow, low-rate drift that preserves short-term consistency.}

In this paper, we present a lightweight, unsupervised, headset-side defense for edge-offloaded VIO. 
Our approach models the natural consistency between server-returned “slow” poses and locally integrated “fast” poses. We design a deep autoencoder, trained only on clean pose and IMU data from attack-free runs, that learns intrinsic motion regularities. 
At runtime, each incoming slow pose is evaluated using the autoencoder’s reconstruction error over a window of preceding fast-pose and IMU features. Our simple policy design accepts normal poses, drops anomalies, or forces a pass after many consecutive drops to re-anchor the fast-pose stream and avoid its own drift.
We evaluate our system on an ILLIXR-based testbed~\cite{illixr}, under multiple spoofing configurations. We use clean-run trajectories as reference ground truth for offline Absolute Trajectory Error (ATE) and Relative Pose Error (RPE). The results show that under moderate spoofing, the defense improves both metrics by more than 10x with minimal added latency, on the order of a few milliseconds per slow-pose frame on a typical edge server.

The remainder of the paper is organized as follows. Section~\ref{sec:sys-model} details the problem formulation. Section~\ref{sec:solution} presents the defense pipeline. Section~\ref{sec:evaluation} reports the experimental results. Section~\ref{sec:conclusion} concludes the paper.

\section{Problem Formulation}
\label{sec:sys-model}

\begin{figure}[t]
  \centering
  \includegraphics[width=0.9\columnwidth]{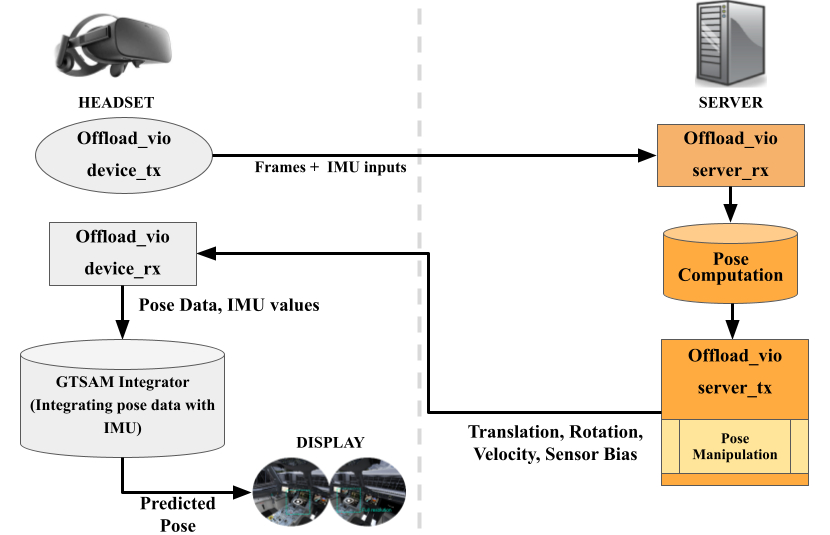}
  \caption{Headset offloads sensors inputs to edge and computes local fast-pose window; edge server returns slow-pose.}
  \label{fig:setup}
\end{figure}

\subsection{System Model}


We build on the ILLIXR framework and the RemoteVIO paradigm \cite{10.1145/3712676.3714442} by simulating the VR client on a PC using the ILLIXR application, that streams prerecorded image and IMU data to an edge server for full VIO computation, while retaining on‐device local pose computation to preserve responsiveness.

As shown in Figure~\ref{fig:setup}, on the client, the \texttt{device\_tx} plugin packages the latest image frames and IMU readings into a Protocol Buffer message and sends it to the server via TCP. While awaiting the server response, the headset integrates IMU data to maintain a local \emph{fast-head-pose buffer} $\mathcal{F}_i = \{\mathbf{f}_{i,1}, \dots, \mathbf{f}_{i,K}\}$, ensuring real-time visual feedback.

Upon receiving the slow head pose $\mathbf{s}_i$, the \texttt{device\_rx} plugin deserializes it and uses it to re-anchor local fast-pose integration for the subsequent window $\mathcal{F}_{i+1}$. This prevents drift from propagating across windows and keeps the client trajectory aligned with the server-computed global pose.

At the server, the \texttt{server\_rx} plugin deserializes the sensor data and runs a full VIO pipeline (OpenVINS) to estimate $\mathbf{s}_i$:
{\small
\begin{equation}
\mathbf{s}_i =
\left[ \mathbf{p}_i, \mathbf{q}_i, \mathbf{v}_i,
\mathbf{b}_i^{\text{acc}}, \mathbf{b}_i^{\text{gyro}} \right],
\end{equation}
}

where $\mathbf{p}_i$ is 3D position, $\mathbf{q}_i$ the orientation quaternion, $\mathbf{v}_i$ the linear velocity, and $\mathbf{b}_i^{\text{acc}}, \mathbf{b}_i^{\text{gyro}}$ the accelerometer and gyroscope biases. The \texttt{server\_tx} plugin serializes and returns $\mathbf{s}_i$ to the client, completing the round and re-aligning local integration with the global server trajectory.
gration with the global server trajectory.

\subsection{Threat Model}
\label{sec:threat}

We assume the edge server that executes the offloaded VIO pipeline is adversary-controlled (Figure~\ref{fig:setup}). At each offload round \(i\), the adversary spoofs the returned slow pose with probability \(p\); let \(b_i \sim \mathrm{Bernoulli}(p)\) denote this event. If \(b_i=1\), the true slow-pose state \(\mathbf{s}_i=[\mathbf{p}_i,\mathbf{q}_i,\mathbf{v}_i,\mathbf{b}_i]\), with \(\mathbf{b}_i=[\mathbf{b}_i^{\text{acc}};\mathbf{b}_i^{\text{gyro}}]\), is perturbed by small, bounded additive drifts \(\delta_i=[\delta_i^{\text{pos}},\delta_i^{\text{ang}},\delta_i^{\text{vel}},\delta_i^{\text{bias}}]\) to produce \(\tilde{\mathbf{s}}_i=\mathbf{s}_i+\delta_i\); otherwise (\(b_i=0\)) the pose is returned unmodified. The client does not observe \(b_i\) and has no ground-truth labels or information to distinguish spoofed from clean rounds. Its inputs at round \(i\) are the possibly spoofed slow pose \(\tilde{\mathbf{s}}_i\) and the locally integrated fast-pose buffer \(\mathcal{F}_{i-1}\) accumulated since the previous slow-pose update. 
The manipulation is injected after server-side pose computation and before the transmission back to the headset. 

\subsection{Problem Evidence Analysis}

Edge-offloaded VIO pipelines inherently trust slow-pose updates from the server, with no access to ground truth or cues about potential perturbations. Small, consistent drifts can silently corrupt the motion trajectory, accumulating over time without triggering residual filters or anomaly thresholds.

\begin{figure}[t]
    \centering
    \begin{subfigure}[t]{0.45\linewidth}
        \includegraphics[width=\linewidth]{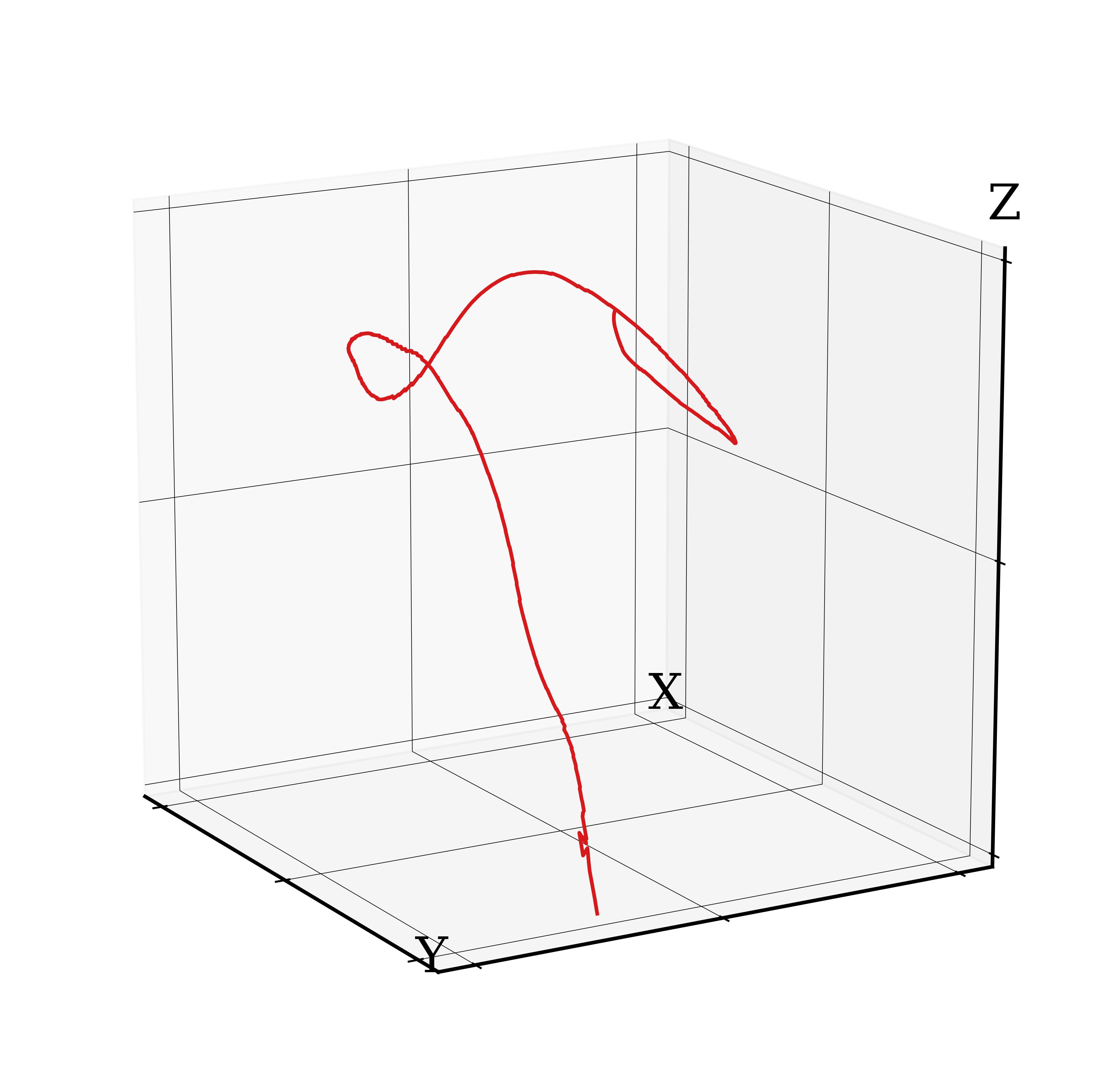}
        \caption{No spoofing (0\%)}
        \label{fig:traj_nochange}
    \end{subfigure}
    \begin{subfigure}[t]{0.45\linewidth}
        \includegraphics[width=\linewidth]{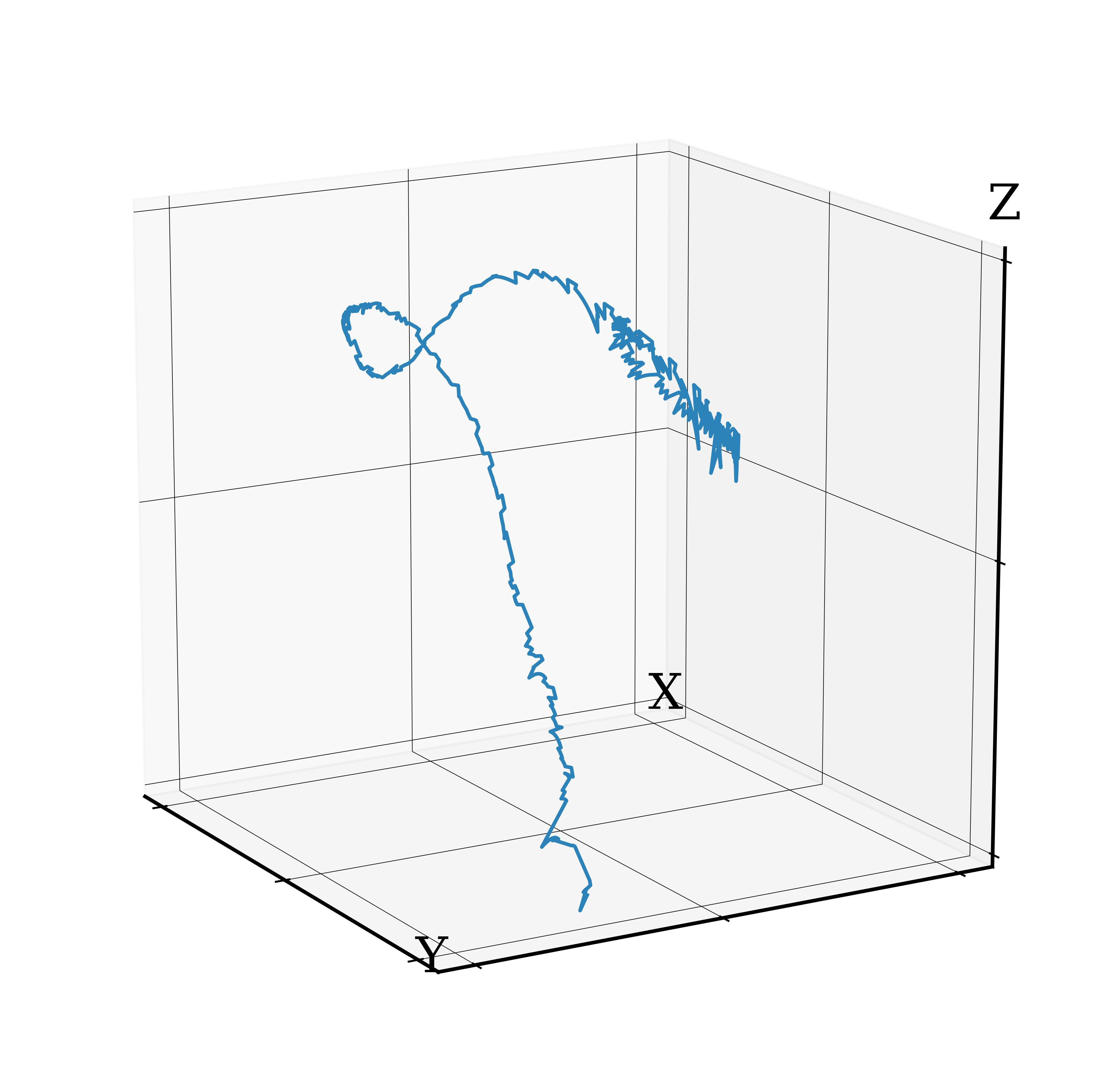}
        \caption{Significant spoofing (75\%)}
        \label{fig:traj_25}
    \end{subfigure}
    \caption{Client 3D trajectories: normal vs. adversarial spoofing; Subplot \subref{fig:traj_25} shows spoofing-induced drift.}
    \label{fig:gtsam_traj_all}
     \vspace{-5pt}
\end{figure}

Each slow pose anchors the next fast-pose integration window, and when spoofed, errors propagate downstream, compounding drift in fast poses. As depicted in Figure~\ref{fig:gtsam_traj_all}, trajectories diverge and exhibit jitter with the onset of spoofing, which worsens with increased spoofing probability; even severe drift affects slow poses, degrading spatial fidelity. These errors compromise scene stability, creating jitter and user fatigue. Given the stealthy nature of perturbations and the lack of client verification capabilities, a real-time detection and correction mechanism on the headset is crucial to preserving motion accuracy and QoE in immersive XR.

\subsection{Problem Statement}

Given input pairs \(\{(\tilde{\mathbf{s}}_i, \mathcal{F}_{i-1})\}_{i=1}^N\), where each slow pose \(\tilde{\mathbf{s}}_i\) may be adversarially perturbed, the client must:

\begin{itemize}
  \item \textbf{Detection:} Identify which slow poses are anomalous by evaluating their consistency with preceding motion.
  \item \textbf{Decision:} Choose whether to accept \(\tilde{\mathbf{s}}_i\) or drop it, thereby influencing how the subsequent fast poses are integrated.
\end{itemize}

We evaluate the effectiveness of this decision process using the headset’s \textit{fast pose outputs} 
$\{\mathbf{f}_t^{\text{out}}\}_{t=1}^T$, where $T \gg N$ denotes the total number of high-rate fast-pose 
estimates produced during runtime. The evaluation metrics are:

{\small
\begin{equation}
\mathrm{ATE} = \sqrt{\tfrac{1}{T} \sum_{t=1}^T \left\| \mathbf{f}_t^{\text{out}} - \mathbf{f}_t^{\text{gt}} \right\|^2},
\quad
\mathrm{RPE} = \sqrt{\tfrac{1}{T-1} \sum_{t=1}^{T-1} \left\| \Delta\mathbf{f}_t^{\text{out}} - \Delta\mathbf{f}_t^{\text{gt}} \right\|^2},
\label{eq:ate_rpe}
\end{equation}
}

where $\mathbf{f}_t^{\text{gt}}$ denotes the ground-truth pose at timestamp $t$, and 
$\Delta\mathbf{f}_t = \mathbf{f}_{t+1} \ominus \mathbf{f}_t$ represents the relative motion 
(translation or rotation). The goal is to ensure that corrupted slow poses do not degrade the 
downstream fast-pose trajectory.

\section{Defense Framework}
\label{sec:solution}

We propose a lightweight, client-side framework to detect and mitigate adversarial drift in slow pose estimates using an unsupervised approach. The system leverages the temporal consistency between fast and slow pose streams and operates without labeled spoofed data, making it practical for real-time deployment on XR headsets.

Since the client has no access to ground-truth drift labels or attacker strategy, detection must rely solely on motion patterns learned from clean, attack-free runs. This is viable because pose estimation in XR is a deterministic function of sensor input: for a given motion, the resulting pose trajectory is governed by sensor fusion and is independent of the XR application in use. Thus, clean motion windows recorded under normal conditions generalize across applications, enabling an autoencoder trained on these sequences to detect deviations introduced by spoofed poses.

\subsection{Feature Extraction and Temporal Encoding}

For each slow pose \(s_i\), we form features from the prior window between the last accepted \(s_{i-1}\) and \(s_i\)—fast poses \(\mathcal{F}_{i-1}=\{\mathbf{f}_{i-1,k}\}_{k=1}^{K}\) and IMU samples \(\mathcal{I}_{i-1}=\{\mathbf{z}_{i-1,k}\}_{k=1}^{K}\)—to avoid contamination from spoofed \(s_i\). The vector includes window stats (count \(K\), duration); pose residuals between \(\mathbf{f}_{i-1,k}\) and \(s_i\) for position/velocity summarized by mean/std/min/max and \(\ell_1\) norm; quaternion geodesic orientation error; accelerometer/gyroscope bias summaries; and IMU activity (norms and axis-wise statistics). Features are z-normalized, PCA-reduced to 97\% variance, and fed to the autoencoder for anomaly scoring.

\begin{figure}[!t]
\centering
\includegraphics[width=\linewidth]{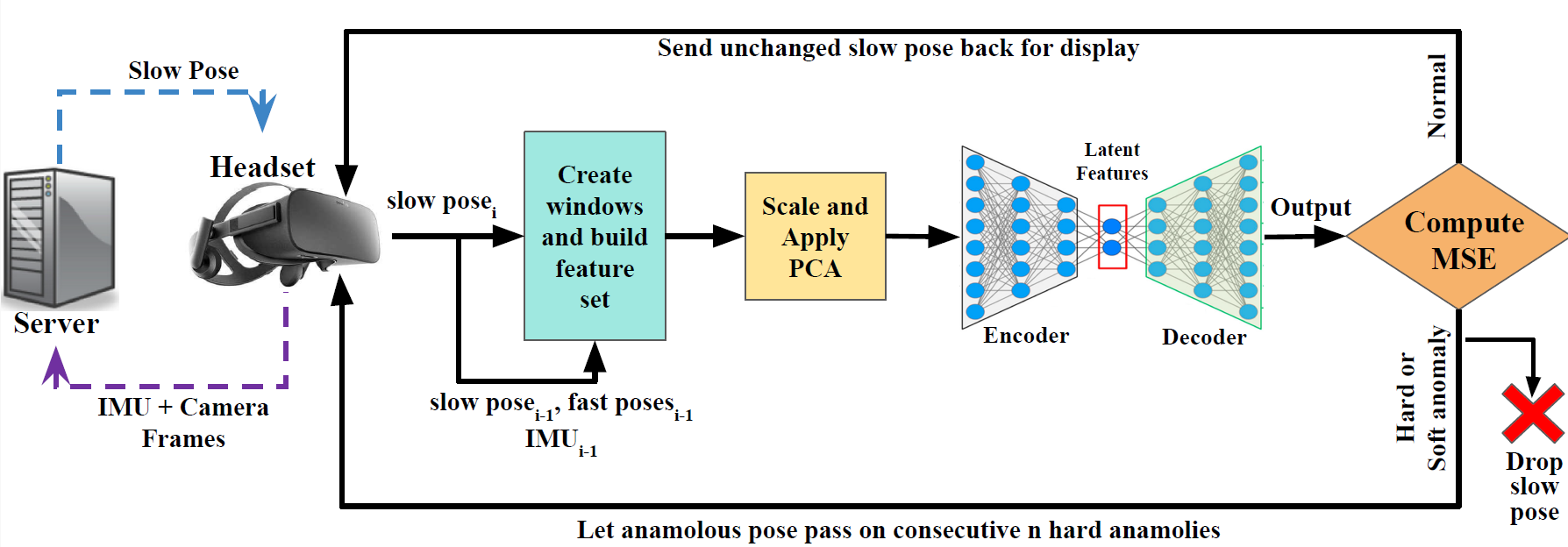}
\caption{Runtime pipeline: features for each slow pose constructed using preceding fast-pose/IMU window; reconstruction error decides accept/drop/force-pass.}
\label{fig:system_diagram}
\end{figure}

\subsection{Autoencoder-Based Anomaly Detection}
Figure ~\ref{fig:system_diagram} illustrates the defense pipeline. We train a compact fully connected autoencoder on clean logs to reconstruct PCA-reduced features summarizing the fast-pose/IMU window preceding each slow pose (architecture in Table~\ref{tab:ae_architecture}). The encoder maps inputs to a low-dimensional latent and the decoder reconstructs them; training uses SmoothL1 (Huber) loss with light Gaussian jitter and early stopping. At runtime, the same feature/PCA steps feed the model and the reconstruction mean-squared error (MSE) serves as the anomaly score; spoofed inputs typically yield larger errors. Thresholds calibrated on a clean validation set define 3 anomaly types: \textit{Normal} when MSE is below median + 3$\times$ the median absolute deviation (MAD), \textit{Soft anomaly} when the MSE lies between that value and the 98th percentile, and \textit{Hard anomaly} when at or above the 98th percentile. Normal poses are accepted; soft and hard anomalies are dropped. To avoid prolonged drift during extended hard anomaly drops, we allow a forced pass after 12 consecutive hard classifications to re-anchor the fast-pose stream.

\begin{table}[htbp]
\caption{Proposed autoencoder architecture.}
\label{tab:ae_architecture}
\centering
\scriptsize
\begin{tabular}{@{}llll@{}}
\toprule
\textbf{Stage} & \textbf{Layer} & \textbf{Output Size} & \textbf{Activation / Notes} \\
\midrule
\multirow{4}{*}{Encoder} 
  & Linear (input $\rightarrow$ 256) & 256 & LReLU + BN1d \\
  & Linear (256 $\rightarrow$ 128)   & 128 & LReLU + BN1d \\
  & Linear (128 $\rightarrow$ 64)    & 64  & LReLU + BN1d \\
  & Linear (64 $\rightarrow$ LATENT\_DIM) & 32 & -- \\
\midrule
\multirow{4}{*}{Decoder} 
  & Linear (LATENT\_DIM $\rightarrow$ 64) & 64  & LReLU \\
  & Linear (64 $\rightarrow$ 128)         & 128 & LReLU \\
  & Linear (128 $\rightarrow$ 256)        & 256 & LReLU \\
  & Linear (256 $\rightarrow$ input dim)  & input size & Tanh \\
\bottomrule
\end{tabular}
\normalsize
\vspace{-0.6em} 
\end{table}

\section{Evaluation}
\label{sec:evaluation}
\subsection{Experimental Setup}

We evaluate the framework in real time on ILLIXR using two PCs (Intel Core i9-14900K, NVIDIA RTX 2000, 32~GB RAM): the client emulates the headset and the server runs VIO. The server’s VIO plugin emits clean and spoofed slow poses, with drift magnitudes and Bernoulli injection probability \(p\) set via a configuration file to sweep attack intensities. For autoencoder training, we collect client-side IMU, fast poses, and server-returned slow poses across 100 clean runs, logging per-run poses (position, velocity, quaternion), sensor biases, and raw IMU traces. On the client, a lightweight Python sidecar receives fast poses/IMU over ZMQ and slow poses over TCP, builds per–slow-pose features from the preceding fast-pose/IMU window, applies normalization+PCA, and scores anomalies with the trained autoencoder; the resulting decision (accept, soft/hard reject, or forced pass) is returned to ILLIXR to anchor the next fast-pose window. All detection/decision logic remains external to ILLIXR with minimal hooks. All 100 clean-run logs, training code, and the runtime detection-and-recovery pipeline are available at \url{https://github.com/Sourya17/VR_PoseDriftRecovery}.

\subsection{Metrics}

We evaluate our framework with three complementary measures that capture global drift, local motion consistency, and detector fidelity: Absolute Trajectory Error (ATE)—Euclidean distance between estimated and reference trajectories; Relative Pose Error (RPE)—deviation in frame-to-frame motion; and autoencoder reconstruction error (MSE)—mean-squared difference between the PCA-reduced feature input and its reconstruction. We do not use native (non-offloaded) ILLIXR ground truth as the ATE/RPE reference because even small, systematic trajectory drift between native and offloaded runs would overshadow the subtler effects of spoofing and the defense. Instead, clean offloaded trajectories collected under identical conditions serve as the reference, isolating the impact of perturbations and the defense.

\subsection{Results}

Unless otherwise noted, all results in Sections~\ref{sec:result_first}–\ref{sec:result_fourth} use a fixed set of attacker parameters to inject low-magnitude drifts into slow-pose updates. The spoofing was controlled via environment variables on the server in Table \ref{tab:spoof_env_vars}.

\begin{table}[htbp]
\caption{Server environment variables.}
\label{tab:spoof_env_vars}
\centering
\scriptsize
\begin{tabular}{l l}
\toprule
\textbf{Variable} & \textbf{Value} \\
\midrule
\texttt{ILLIXR\_VIO\_SPOOF\_BIAS\_DRIFT}     & 0.05 \\
\texttt{ILLIXR\_VIO\_SPOOF\_VELOCITY\_DRIFT} & 0.10 \\
\texttt{ILLIXR\_VIO\_SPOOF\_POSITION\_DRIFT} & 0.02 \\
\texttt{ILLIXR\_VIO\_SPOOF\_ANGLE\_DRIFT}    & 0.20 \\
\bottomrule
\end{tabular}
\vspace{-0.1in}
\end{table}

These values were held constant across spoofing rates (0\%, 25\%, 50\%, and 75\%) to isolate the impact of spoof frequency under a uniform attack model.

Section~\ref{sec:result_fifth} explores defense robustness by varying these parameters, testing generalization to unseen attack profiles. For all experiments, we report results from the best of 10 ILLIXR runs per condition, both with and without the defense enabled.

\subsubsection{ATE and RPE without the Defense Strategy}
\label{sec:result_first}

We begin by quantifying the impact of spoofed slow poses on the offloaded VIO pipeline in the absence of any defense. Table~\ref{tab:ate_rpe_means} reports the mean translation and rotation ATE and RPE across spoofing levels of 0\%, 25\%, 50\%, and 75\%. As expected, all metrics remain negligible for the clean 0\% case. However, even moderate spoofing at 25\% causes a noticeable increase (e.g., T-ATE: 10.3~cm, R-RPE: 2.45$^\circ$). As spoofing frequency increases to 50\% and 75\%, the degradation compounds further, with rotational errors rising to 2.98$^\circ$ and 2.57$^\circ$, respectively.

\begin{table}[htbp]
\caption{Mean ATE and RPE under varying spoofing levels without any defense strategy.}
\label{tab:ate_rpe_means}
\centering
\scriptsize
\begin{tabular}{lrrrr}
\toprule
\textbf{Spoofing Level} & \textbf{T-ATE (cm)} & \textbf{R-ATE (deg)} & \textbf{T-RPE (cm)} & \textbf{R-RPE (deg)} \\
\midrule
0\%  & 0.074   & 0.008   & 0.033  & 0.0019 \\
25\% & 10.319  & 16.353  & 0.815  & 2.453  \\
50\% & 13.043  & 33.904  & 1.449  & 2.980  \\
75\% & 14.336  & 49.465  & 1.709  & 2.567  \\
\bottomrule
\end{tabular}
\vspace{-0.2in}
\end{table}

\begin{figure}[htbp]
  \centering
  \begin{subfigure}[t]{0.8\linewidth}
    \centering
    \includegraphics[width=\linewidth]{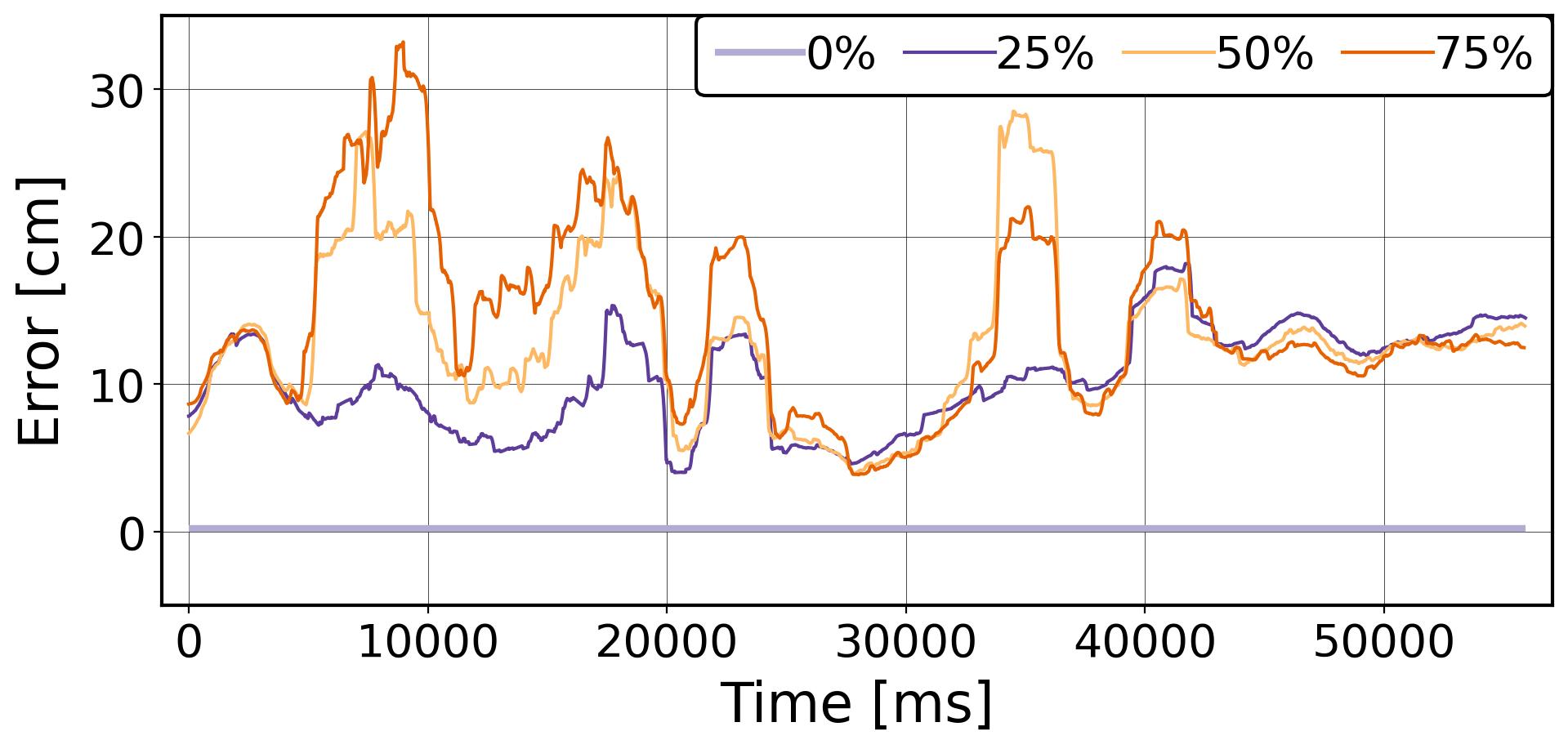}
    \caption{Smoothed translation ATE over time} 
    \label{subfig:ate_time}
  \end{subfigure}

  \vspace{0.6em}

  \begin{subfigure}[t]{0.8\linewidth}
    \centering
    \includegraphics[width=\linewidth]{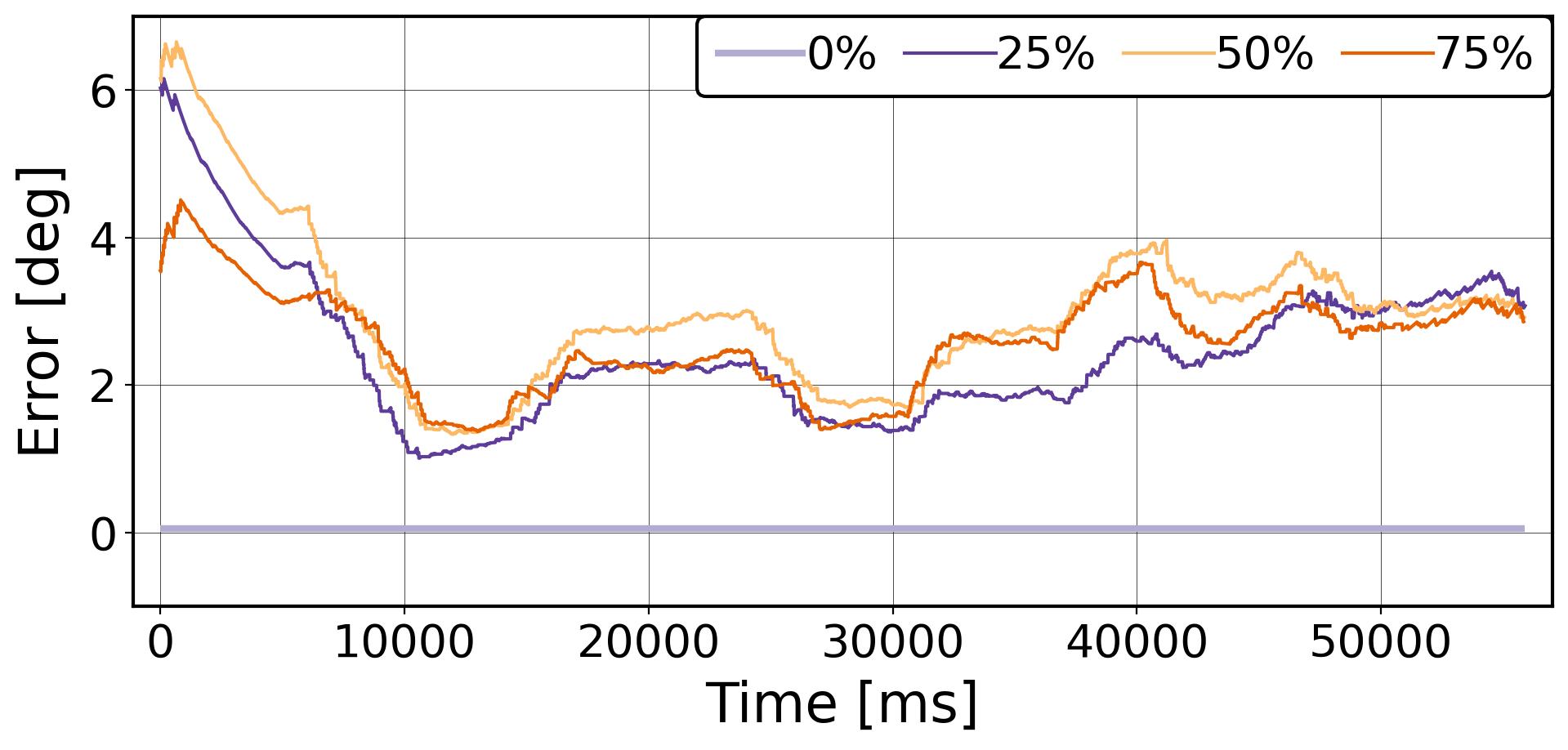}
    \caption{Smoothed rotation RPE over time} 
    \label{subfig:rpe_time}
  \end{subfigure}

  \caption{Evaluation of pose degradation without defense under different spoofing levels. Higher spoofing levels lead to increasingly persistent and severe trajectory errors.}
  \label{fig:pose_error_wo_defense}
\end{figure}

Figure~\ref{fig:pose_error_wo_defense} presents the temporal trends of translation ATE and rotation RPE. The clean run shows flat error curves, whereas higher spoofing levels exhibit growing plateaus of drift. Notably, the 75\% spoofed run reveals extended regions of high trajectory error, showing how compounded spoofed poses corrupt fast-pose propagation over time.
These results highlight the vulnerability of offloaded VIO to persistent low-magnitude spoofing and establish the need for robust real-time anomaly detection on the client side.

\subsubsection{Autoencoder Performance}
\label{sec:result_second}

\begin{figure}[t]
  \centering
  \begin{subfigure}[t]{0.48\linewidth}
    \centering
    \includegraphics[width=\linewidth]{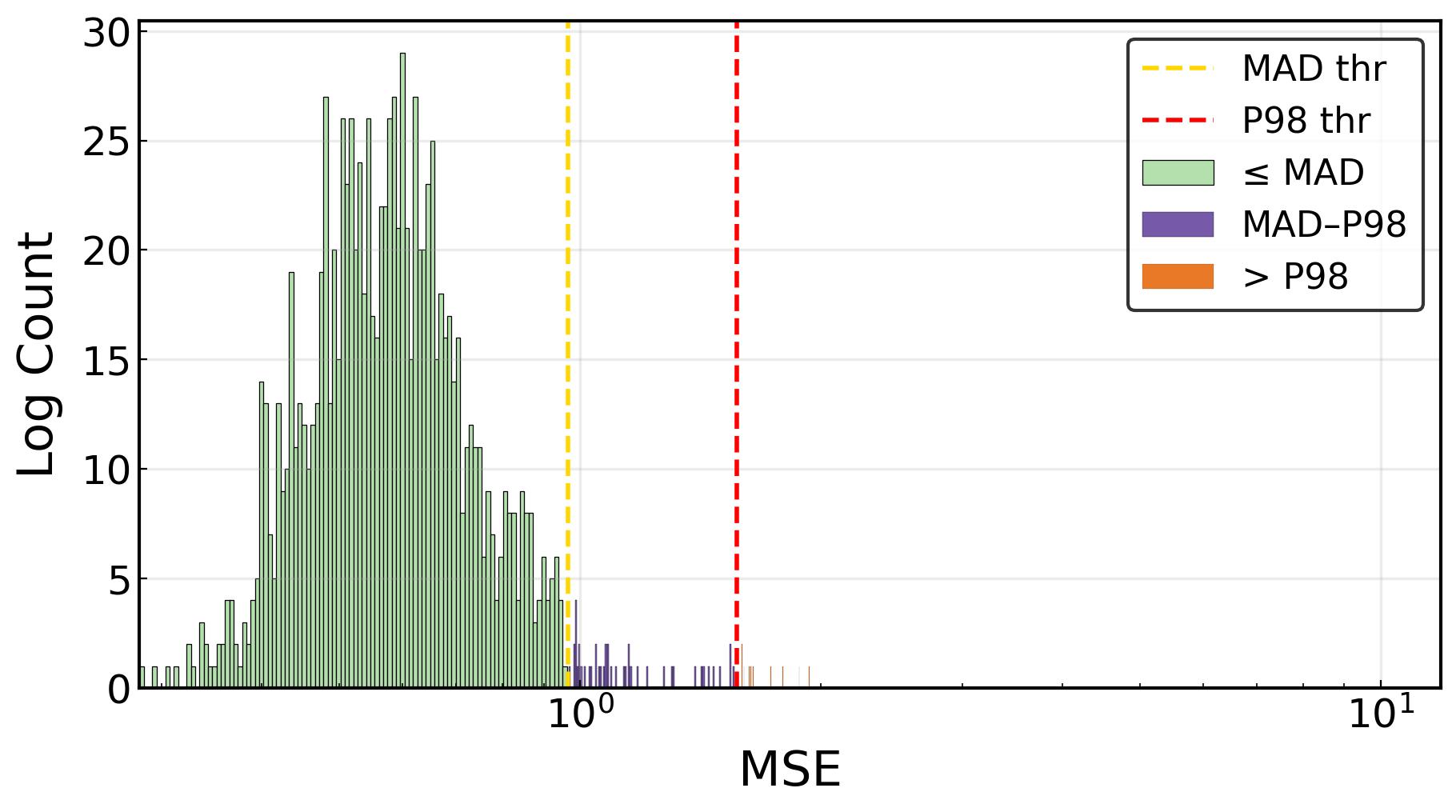}
    \caption{No spoofing (0\%)}
  \end{subfigure}
  \hfill
  \begin{subfigure}[t]{0.48\linewidth}
    \centering
    \includegraphics[width=\linewidth]{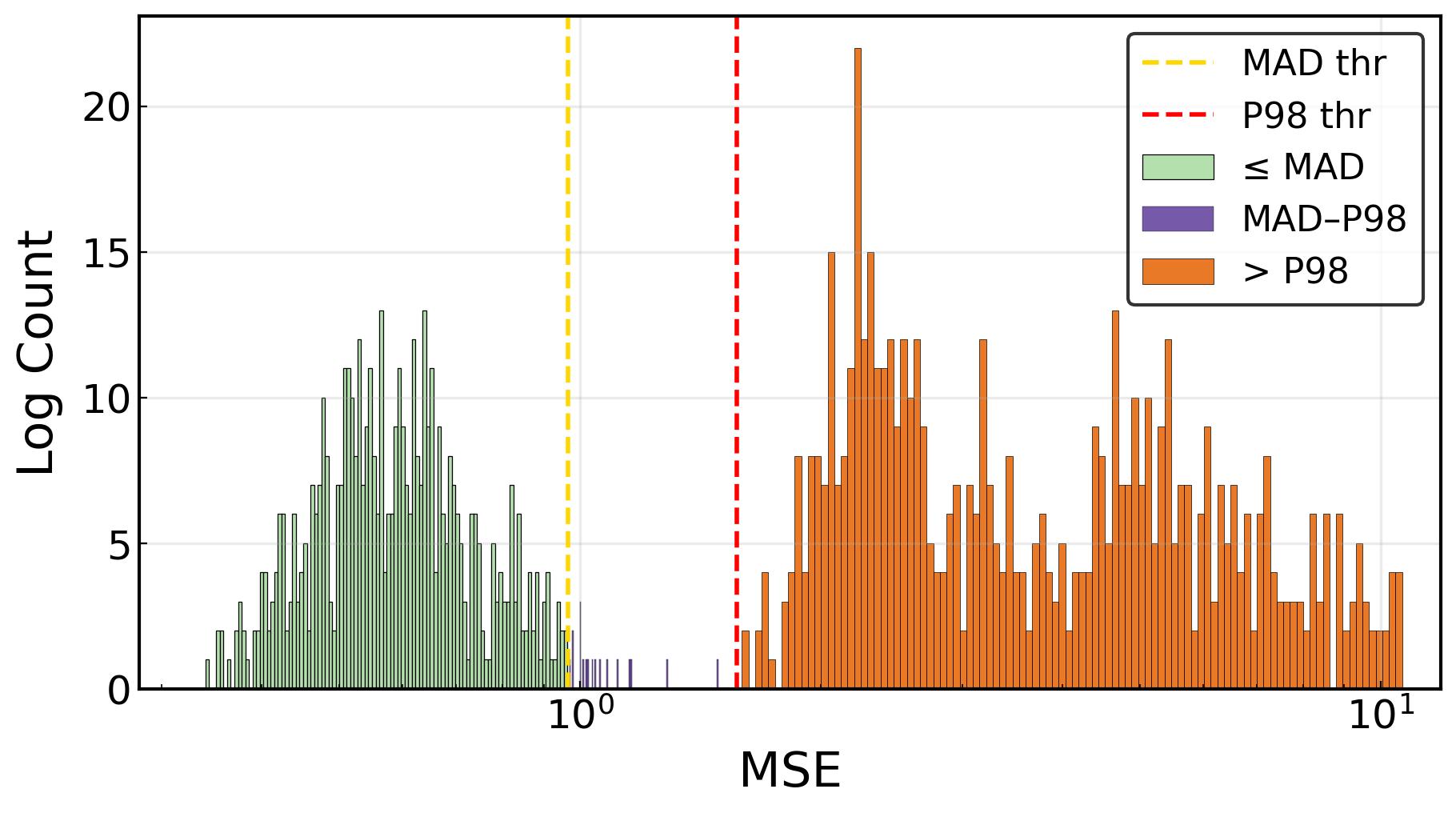}
    \caption{25\% spoofing}
  \end{subfigure}
  \vskip\baselineskip
  \vspace{-0.15in}
  \begin{subfigure}[t]{0.48\linewidth}
    \centering
    \includegraphics[width=\linewidth]{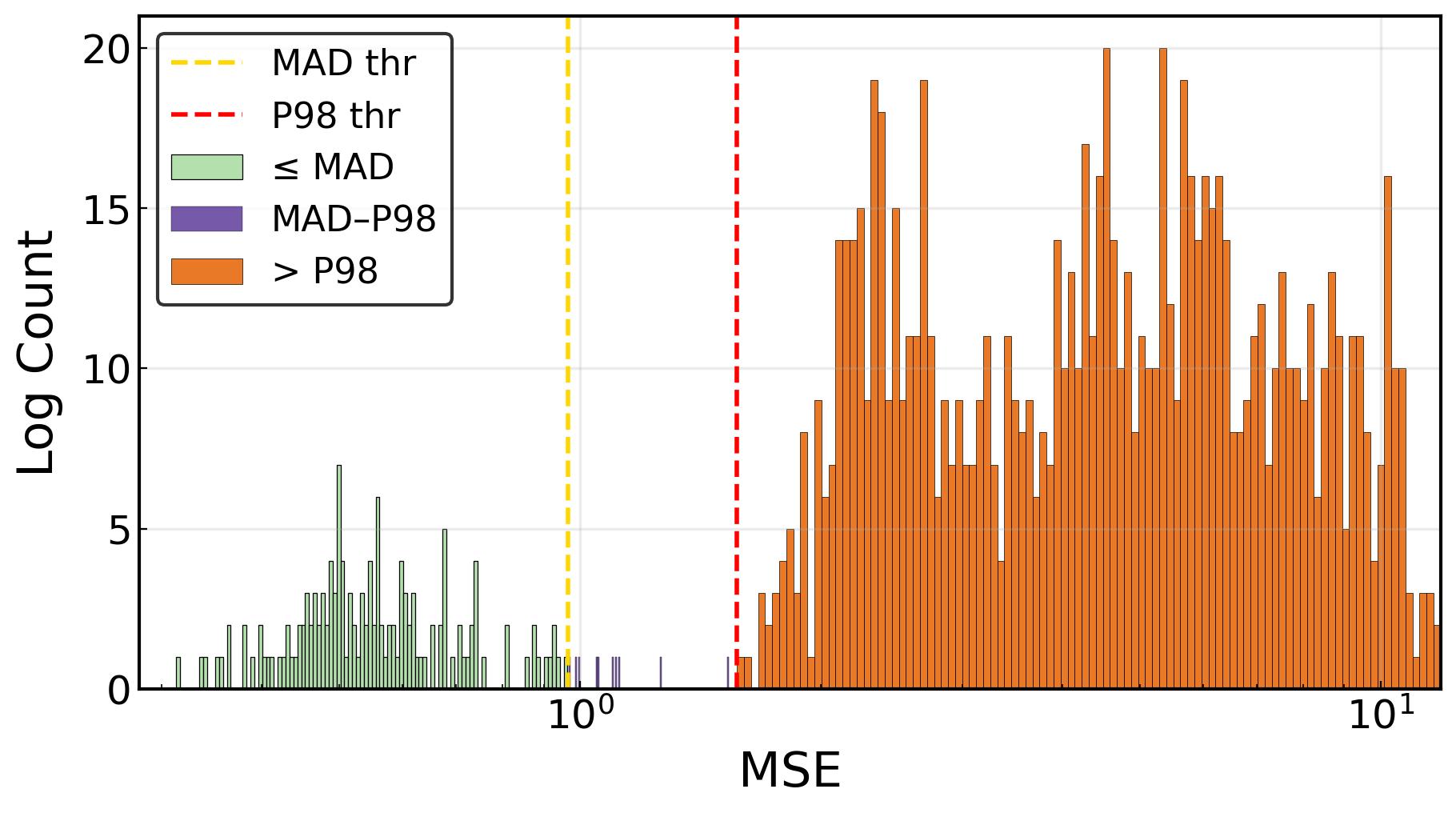}
    \caption{50\% spoofing}
  \end{subfigure}
  \hfill
  \begin{subfigure}[t]{0.48\linewidth}
    \centering
    \includegraphics[width=\linewidth]{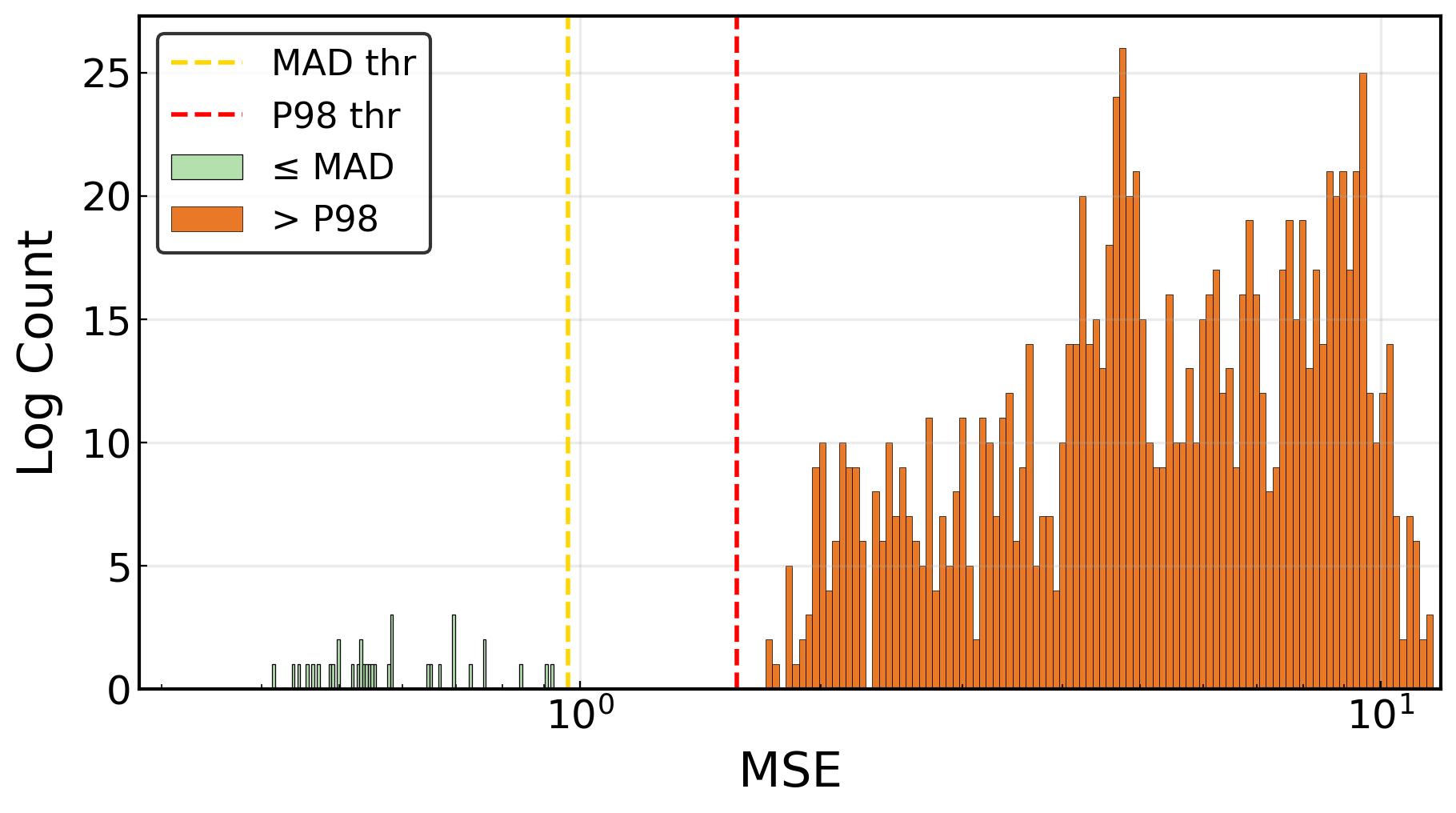}
    \caption{75\% spoofing}
  \end{subfigure}
  \caption{Histogram of autoencoder reconstruction errors (MSE) across four spoofing scenarios. The yellow line denotes the MAD-based threshold; the red line shows the 98th percentile. As spoofing increases, MSE distribution shifts right.}
  \label{fig:mse_grid}
\end{figure}

We analyze the autoencoder’s reconstruction error (MSE) across four spoofing levels—0\%, 25\%, 50\%, and 75\%—with results shown in Figure~\ref{fig:mse_grid}. Each histogram plots per-frame MSE over 100 bins, along with the MAD-based anomaly threshold (yellow dashed line) and the 98th percentile threshold (red dashed line).

In the clean setting, MSE values remain tightly clustered well below the threshold, with over 95\% of slow poses classified as normal. As spoofing increases, the distribution shifts rightward. At 25\%, only 44\% of slow poses remain normal, while 54\% are hard anomalies. This trend intensifies at 50\% and 75\%, where 86.7\% and 96.8\% of poses are flagged as hard anomalies.

Interestingly, the fraction of detected anomalies exceeds the spoofing probability. This amplification stems from corrupted slow poses contaminating downstream fast-pose windows, leading to broader temporal inconsistency detectable by the autoencoder—even if the corruption was sparse. Despite this, the response remains monotonic: higher spoofing leads to more detections, while the clean case avoids false positives. This confirms the autoencoder’s sensitivity and generalization capability under varied attack intensities.

\subsubsection{Autoencoder Behavior in Passive Detection Mode}
\label{sec:result_third}

\begin{figure}[t]
    \centering
    \begin{subfigure}[t]{0.9\linewidth}
        \centering
        \includegraphics[width=\linewidth]{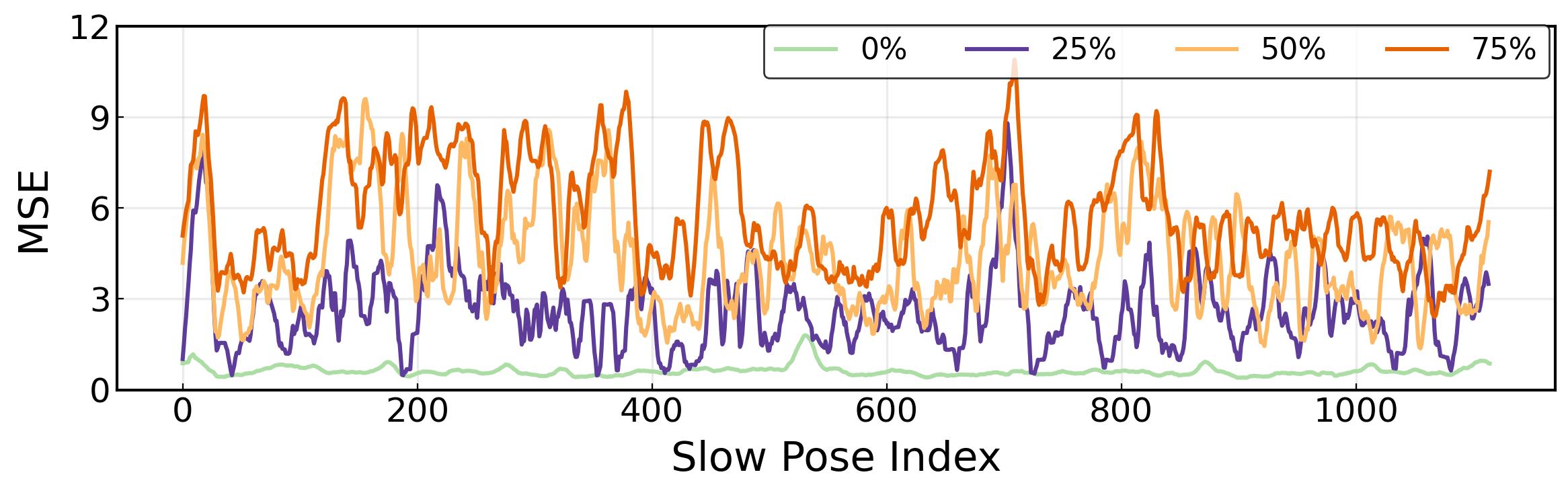}
        \caption{Passive detection mode}
        \label{fig:mse_time_series_passive}
    \end{subfigure}
    
    \vspace{1em} 

    \begin{subfigure}[t]{0.9\linewidth}
        \centering
        \includegraphics[width=\linewidth]{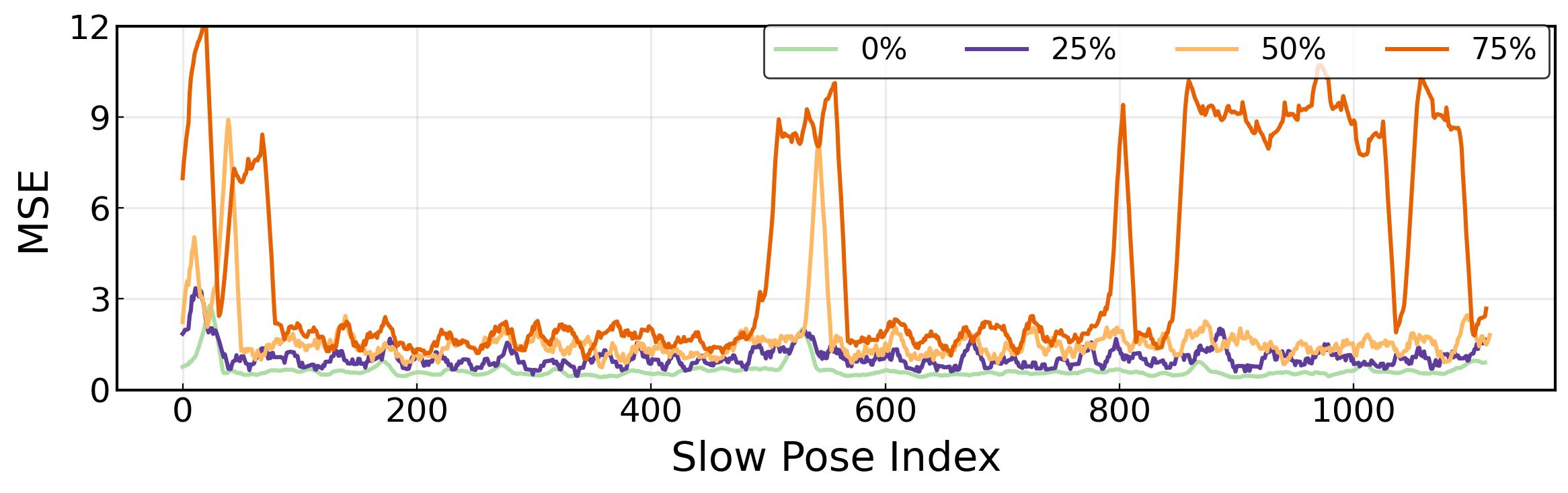}
        \caption{Active detection and recovery mode}
        \label{fig:mse-defense}
    \end{subfigure}

  \caption{MSE over time (averaged over 100 slow-pose bins) for different spoofing frequencies, where higher spoofing causes larger and more frequent error spikes.}

    \label{fig:mse-combined}
\end{figure}

To isolate detector behavior from decision-making, we run full real-time feature extraction and autoencoder inference on the client but forward all slow poses unchanged. Figure~\ref{fig:mse_time_series_passive} shows MSE traces for spoofing rates of 0\%, 25\%, 50\%, and 75\%. Clean runs remain low and stable, while increasing spoofing produces progressively higher and denser spikes, culminating in near-continuous peaks at 75\%—indicating complete breakdown of fast–slow correlation. This confirms that reconstruction error alone reliably captures spoofing-induced motion inconsistency.

\subsubsection{Anomaly Detection Behavior (MSE Time Series)}
\label{sec:result_fourth}

With the full pipeline active, each slow pose is scored in real time and classified as normal, soft anomaly (drop), or hard anomaly (normally drop, forced pass after a number of subsequent hard anomalies). Figure~\ref{fig:mse-defense} shows that at 0\% spoofing, MSE stays low, avoiding false positives. As spoofing increases, spikes occur but are shorter and less persistent than in passive mode (Section~\ref{sec:result_third}). At high spoofing (75\%), periodic forced passes cause brief MSE surges, but the system quickly re-stabilizes, preventing cumulative drift. Overall, active defense both limits spike duration and suppresses downstream error growth under sustained attack.

\subsubsection{Pose Quality Under Full Defense}

We now evaluate the full defense system’s ability to suppress trajectory degradation under increasing spoofing levels. Table~\ref{tab:ate_rpe_means_defense} reports mean translational and rotational ATE and RPE for spoofing rates of 0\%, 25\%, 50\%, and 75\%. Compared to the unprotected case (Table~\ref{tab:ate_rpe_means}), our defense substantially reduces error in almost all metrics across all spoofing intensities.

\begin{table}[ht]
  \caption{Mean ATE and RPE under varying spoofing levels} 
  \label{tab:ate_rpe_means_defense}
  \scriptsize
  \centering
  \begin{tabular}{lrrrr}
    \toprule
    \textbf{Spoofing} & \textbf{T-ATE} & \textbf{R-ATE} & \textbf{T-RPE} & \textbf{R-RPE} \\
    \textbf{Level} & (cm) & (deg) & (cm) & (deg) \\
    \midrule
    0\,\%  & 0.400 & 0.049 & 0.043 & 0.0036 \\
    25\,\% & 0.388 & 0.049 & 0.044 & 0.0039 \\
    50\,\% & 1.369 & 2.166 & 0.082 & 0.041 \\
    75\,\% & 27.812 & 23.213 & 0.979 & 0.146 \\
    \bottomrule
  \end{tabular}
\end{table}

At 0\% and 25\% spoofing, ATE and RPE remain nearly indistinguishable from clean baseline: all metrics are below 0.5\,cm and 0.05$^\circ$, indicating robust rejection of false positives and stability in benign scenarios. At 50\% spoofing, the defense suppresses translational ATE from over 13\,cm to 1.37\,cm and rotational ATE from 33$^\circ$ to 2.17$^\circ$. Translational and rotational RPE both fall below 0.1, confirming strong correction of local motion consistency.

At 75\%, although RPE remains low (0.98\,cm and 0.15$^\circ$), cumulative ATE rises (27.81\,cm and 23.21$^\circ$). This occurs due to frequent forced passes of anomalous poses that re-anchor fast pose streams with uncorrected values. While this helps preserve scene continuity under severe spoofing, it leads to global drift over time.

\begin{figure}[t]
  \centering
  \begin{subfigure}[t]{0.85\linewidth}
    \centering
    \includegraphics[width=\linewidth]{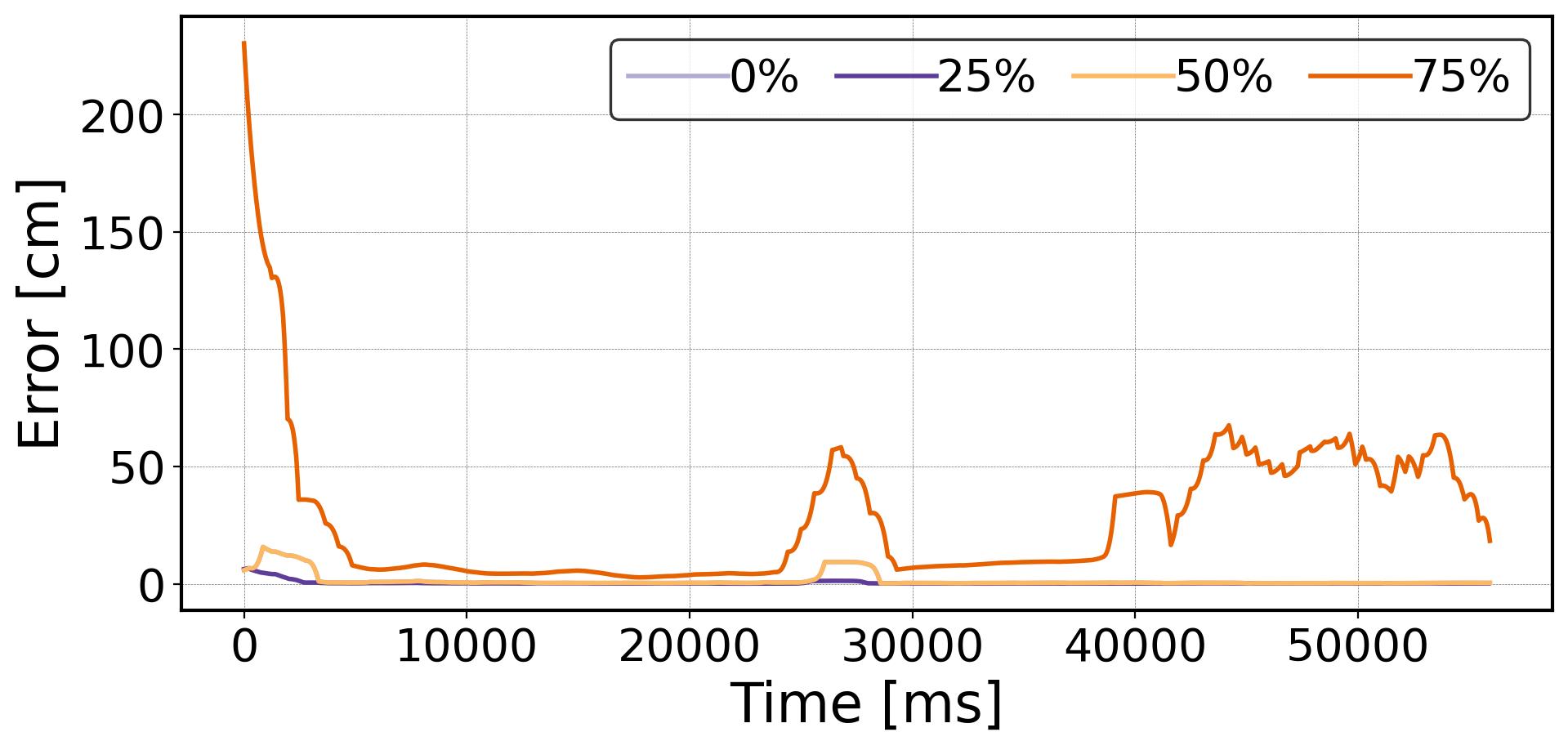}
    \caption{Smoothed translation ATE over time (500-frame window).}
    \label{subfig:ate_time_def}
  \end{subfigure}

  \vspace{0.6em}

  \begin{subfigure}[t]{0.85\linewidth}
    \centering
    \includegraphics[width=\linewidth]{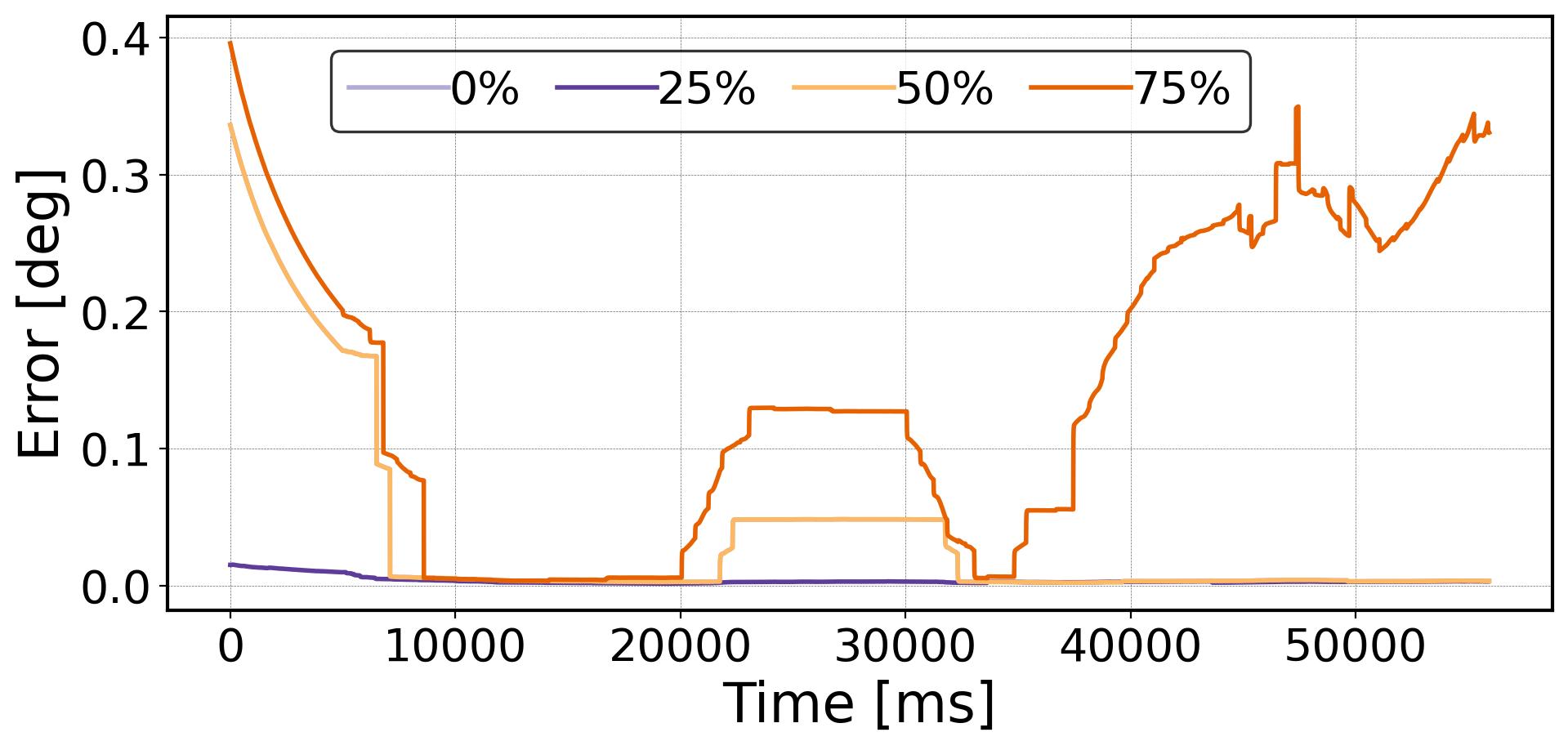}
    \caption{Smoothed rotation RPE over time (2000-frame window).}
    \label{subfig:rpe_time_def}
  \end{subfigure}

  \caption{Evaluation with complete defense under different spoofing levels. Panels show translation ATE (top) and rotation RPE (bottom) over time.}
  \label{fig:pose_error_defense}
\end{figure}

Temporal plots in Figure~\ref{fig:pose_error_defense} (top: ATE, bottom: RPE) further highlight the contrast. Compared to the unprotected case in Figure~\ref{fig:pose_error_wo_defense}, the defense curves are significantly flatter. Translation ATE remains below 10\,cm for 25\% and 50\%, and even the 75\% curve shows rapid recovery after spikes. Rotation RPE stays under 0.4$^\circ$ across all cases—an order-of-magnitude improvement over the 6$^\circ$ peak seen without defense.

\vspace{1em}
\noindent

\subsubsection{Robustness to Varying Attack Configurations}
\label{sec:result_fifth}

To evaluate the generalization ability of our defense, we test it against four \textit{increasingly aggressive} attack configurations (Config~1–4), each defined by different drift magnitudes in bias, velocity, position, and orientation (Table~\ref{tab:attack_config_rpe}). We fix the spoofing probability at 50\%—a balanced choice that allows degradation to surface without fully destabilizing the system. Lower spoofing (25\%) proved too subtle for clear differentiation, while 75\% often led to collapse regardless of configuration.

We focus on RPE instead of ATE as the primary metric. ATE is sensitive to transient drift and may remain elevated even after recovery, whereas RPE better reflects real-time pose stability and local consistency.
Across all four configurations, the system maintains sub-1 RPE values for both translation and rotation—even in Config~4, which features the most severe drift values (e.g., 1.2 bias, 1.5 velocity, 1.0 position, 2.0 angle). This indicates strong generalization of our defense to unseen perturbation levels and attack directions.

\begin{table}[htbp]
\caption{Attack configurations and corresponding mean RPE values (50\% spoofing).}
\label{tab:attack_config_rpe}
\centering
\scriptsize
\setlength{\tabcolsep}{3pt}
\begin{tabular}{%
l
>{\centering\arraybackslash}p{0.12\columnwidth}
>{\centering\arraybackslash}p{0.12\columnwidth}
>{\centering\arraybackslash}p{0.12\columnwidth}
>{\centering\arraybackslash}p{0.12\columnwidth}
>{\centering\arraybackslash}p{0.16\columnwidth}
>{\centering\arraybackslash}p{0.16\columnwidth}}
\toprule
\textbf{Config} &
\textbf{Bias Drift} &
\textbf{Velocity Drift} &
\textbf{Position Drift} &
\textbf{Angle Drift} &
\textbf{Trans RPE Mean (cm)} &
\textbf{Rot RPE Mean (deg)} \\
\midrule
Config 1 & 0.05 & 0.10 & 0.02 & 0.2 & 0.0477 & 0.0045 \\
Config 2 & 0.10 & 0.30 & 0.09 & 0.5 & 0.0582 & 0.0062 \\
Config 3 & 0.60 & 0.80 & 0.50 & 0.9 & 0.3139 & 0.0591 \\
Config 4 & 1.20 & 1.50 & 1.00 & 2.0 & 0.0591 & 0.0065 \\
\bottomrule
\end{tabular}
\end{table}

Each result corresponds to the best of 10 runs. While higher drifts causes sharper divergence at spoofed frames, the system consistently stabilizes shortly afterward. These results confirm the resilience of our defense pipeline under a wide range of adversarial conditions, including strong, previously unseen perturbations.

\subsubsection{Real-Time Inference Under Mobile CPU Constraints}
\label{sec:result_sixth}

To assess deployability on XR headsets, we benchmark our anomaly detection pipeline under conditions approximating the Meta Quest 3s, which features a Snapdragon XR2 Gen 2 SoC with octa-core Kryo CPUs (2.0–3.19\,GHz).

We replicate this on a FABRIC~\cite{fabric-2019} node with 8\,GB RAM, 100\,GB storage, and 8 vCPUs from an AMD EPYC 7532 processor, frequency-locked at 2.4\,GHz. Turbo scaling is disabled for consistency.

The full defense pipeline—including feature extraction, preprocessing (StandardScaler + PCA), TorchScript autoencoder inference, and threshold-based postprocessing—executes over 1,035 slow-pose windows. Model and scaler loading occur only once at startup.

As shown in Table~\ref{tab:cpu_latency_mean}, all stages complete well within real-time constraints. Feature extraction is the most expensive at 4.77\,ms per frame, while preprocessing, inference, and postprocessing each require under 0.3\,ms. 
The one-time model load (37.4\,ms) occurs at headset startup, before VIO offloading, and plays no role in per-frame decisions or inference.

\begin{table}[htbp]
\caption{Mean latency per detection stage (in milliseconds), averaged over 1,035 slow pose windows. Model and scaler load time occurs once at startup.}
\label{tab:cpu_latency_mean}
\centering
\scriptsize
\begin{tabular}{l c}
\toprule
\textbf{Stage} & \textbf{Mean Time (ms)} \\
\midrule
Model + Scaler Load (once) & 37.40 \\
Feature Extraction          & 4.77  \\
Preprocessing + PCA         & 0.23  \\
Autoencoder Inference       & 0.26  \\
Threshold Postprocessing    & 0.05  \\
\bottomrule
\end{tabular}
\end{table}

\section{Conclusions}
\label{sec:conclusion}

In this paper, we present a real-time, headset-side defense for adversarial pose spoofing in offloaded XR systems. The proposed lightweight autoencoder-based pipeline monitored incoming slow poses using temporal features from fast-pose and IMU windows, flagging anomalies based on reconstruction error and deciding whether to accept, reject, or forward each pose.
Evaluated on the ILLIXR platform, our method improved ATE and RPE by over 10$\times$ at moderate spoofing levels (50\%), while maintaining low latency on mobile-class CPU hardware. Even under high spoofing (75\%), the defense preserved frame-to-frame consistency and prevents jitter, though partial drift did occur due to forced passes of corrupt poses. Latency benchmarks confirmed deployability on XR-class hardware where the full pipeline ran in real time, with the slowest stage (feature extraction) taking under 5\,ms per slow pose.

Moving forward, we plan to explore reinforcement learning-based strategies for visual degradation recovery, allowing the system to dynamically adapt to persistent or evolving spoofing patterns. 

\bibliographystyle{ACM-Reference-Format}
\bibliography{references}

\end{document}